\documentclass[letterpaper,journal]{IEEEtran}
\usepackage{amsmath,amssymb,amsthm}
\usepackage{graphicx}
\usepackage[final]{microtype}
\usepackage{booktabs}
\usepackage{multirow}
\usepackage{cite}
\usepackage[hidelinks]{hyperref}
\usepackage{xcolor}

\newcommand{\hwArmCimu}{40}

\newcommand{\hwArmLower}{25}

\newcommand{\hwArmN}{42}
\newcommand{\hwArmOurs}{43}
\newcommand{\hwErrCimu}{8.0}
\newcommand{\hwErrEmo}{8.6}

\newcommand{\hwErrMob}{11.0}
\newcommand{\hwErrOurs}{8.4}
\newcommand{\hwFmax}{45}
\newcommand{\hwFmin}{16}
\newcommand{\hwLatCimu}{34}
\newcommand{\hwLatEmo}{63}
\newcommand{\hwLatMob}{57}
\newcommand{\hwLatOurs}{29}

\newcommand{\hwNPush}{71}
\newcommand{\hwNoiseCimu}{2.3}
\newcommand{\hwNoiseEmo}{2.4}
\newcommand{\hwNoiseMob}{2.2}
\newcommand{\hwNoiseOurs}{2.7}

\newcommand{\repoURL}{\url{https://juunghyeon.github.io/hold-before-you-step/}}
\newcommand{\hwLatPair}{24}
\newcommand{\hwLatEarlier}{60 of 69}
\newcommand{\hwLatP}{7\times10^{-11}}
\newcommand{\hwArmP}{0.087}

\newcommand{\hwHoldOurs}{21}

\newcommand{\hwHoldN}{22}
\newcommand{\hwHoldRange}{26--59}
\newcommand{\hwEntryIQR}{59--144}

\newcommand{\hwEntryLat}{69}

\newcommand{\hwMatchK}{25}
\newcommand{\hwArmMatchN}{40}
\newcommand{\hwArmMatchOurs}{15.7}
\newcommand{\hwArmMatchMob}{18.4}
\newcommand{\hwArmMatchCimu}{18.9}
\newcommand{\hwArmMatchLowerMob}{23}
\newcommand{\hwArmMatchLowerCimu}{34}
\newcommand{\hwArmMatchPMob}{0.048}
\newcommand{\hwArmMatchPCimu}{6\times10^{-6}}
\newcommand{\hwArmLiftOurs}{24.4}
\newcommand{\hwArmLiftMob}{21.7}
\newcommand{\hwMatchDelayOurs}{55}
\newcommand{\hwMatchDelayMob}{66}

\newcommand{\hwAblBlocks}{12}
\newcommand{\hwAblScoreOurs}{69}
\newcommand{\hwAblScoreAbl}{0}
\newcommand{\hwAblP}{5\times10^{-6}}
\newcommand{\hwAblHeldOurs}{42}
\newcommand{\hwAblNOurs}{49}
\newcommand{\hwAblHeldAbl}{0}
\newcommand{\hwAblNAbl}{6}

\newcommand{\hwAblAllNAbl}{24}
\newcommand{\hwAblFailRange}{22--56}
\newcommand{\hwAblOursFailRange}{52--84}
\newcommand{\hwAblLift}{23}

\newcommand{\hwSusNAbl}{23}

\newcommand{\hwSusHeldOurs}{19}
\newcommand{\hwSusNOurs}{38}

\theoremstyle{definition}
\newtheorem{proposition}{Proposition}
\newtheorem{definition}{Definition}
\graphicspath{{figures/}}

\newcommand{\costABDisp}{96}
\newcommand{\costABDispB}{62}
\newcommand{\costABDispO}{1}

\newcommand{\costABPeak}{44}

\newcommand{\costABWork}{69}

\newcommand{\costABZero}{12}

\newcommand{\costAN}{60}

\newcommand{\costAOZero}{59}

\newcommand{\costCBRec}{46}
\newcommand{\costCBSt}{82}

\newcommand{\costCN}{48}

\newcommand{\costCOZero}{48}
\newcommand{\ecSevC}{21.8}
\newcommand{\ecSevI}{21.2}
\newcommand{\ecSevP}{0.2}
\newcommand{\ecThreeLatI}{32}
\newcommand{\ecThreeLatMob}{70}

\newcommand{\ecThreeRmsI}{19.6}
\newcommand{\ecThreeRmsLitMax}{28.2}
\newcommand{\ecThreeRmsLitMin}{21.3}

\newcommand{\estCimuFaR}{0.95}

\newcommand{\estCimuLatR}{0.96}

\newcommand{\estEmoFaPred}{0.96}
\newcommand{\estEmoFaR}{0.85}

\newcommand{\estMaxDev}{0.017}

\newcommand{\estMobFaL}{40}
\newcommand{\estMobFaR}{0.85}

\newcommand{\estMobLatL}{30}
\newcommand{\estMobLatR}{0.88}

\newcommand{\estOursFaL}{15}
\newcommand{\estOursFaR}{0.95}

\newcommand{\estOursLatL}{10}
\newcommand{\estOursLatR}{0.97}

\newcommand{\fcDown}{22}
\newcommand{\fcN}{4372}
\newcommand{\fcUp}{40}
\newcommand{\ffCondH}{45}
\newcommand{\ffCondHB}{39}
\newcommand{\ffCondN}{48}
\newcommand{\ffCondSt}{30}
\newcommand{\ffCondStB}{72}

\newcommand{\ffCutB}{49}

\newcommand{\ffCutLi}{29}

\newcommand{\ffH}{65}
\newcommand{\ffHB}{62}
\newcommand{\ffHCg}{48}

\newcommand{\ffHLi}{65}
\newcommand{\ffHMeng}{51}
\newcommand{\ffHMob}{63}

\newcommand{\ffN}{80}

\newcommand{\ffSingB}{45}
\newcommand{\ffSingN}{56}
\newcommand{\ffSingO}{50}
\newcommand{\ffSt}{117}
\newcommand{\ffStB}{228}
\newcommand{\ffStCg}{151}
\newcommand{\ffStLi}{165}
\newcommand{\ffStMeng}{120}
\newcommand{\ffStMob}{181}
\newcommand{\ffSuccB}{17}

\newcommand{\ffSuccN}{24}
\newcommand{\ffSuccO}{15}
\newcommand{\ffZ}{30}
\newcommand{\ffZB}{0}
\newcommand{\ffZCg}{32}

\newcommand{\ffZLi}{13}
\newcommand{\ffZMeng}{32}
\newcommand{\ffZMob}{10}

\newcommand{\jrPB}{<10^{-7}}

\newcommand{\jrSBB}{89}

\newcommand{\jrSOB}{34}

\newcommand{\latMaxDev}{0.03}
\newcommand{\latMaxDiff}{0.024}
\newcommand{\latP}{0.011}

\newcommand{\latZeroF}{20}
\newcommand{\latZeroL}{28}
\newcommand{\lnN}{12}

\newcommand{\lnTwentyCI}{9--53}

\newcommand{\lnZeroCI}{55--95}

\newcommand{\lthP}{0.008}
\newcommand{\lthStCut}{37}

\newcommand{\lthZeroL}{31}
\newcommand{\lthZeroM}{24}

\newcommand{\mcHeldC}{32}
\newcommand{\mcHeldO}{63}
\newcommand{\mcN}{96}
\newcommand{\mcPkC}{94}
\newcommand{\mcPkO}{70}
\newcommand{\mcRecC}{67}
\newcommand{\mcRecO}{85}
\newcommand{\mcStC}{153}
\newcommand{\mcStO}{62}

\newcommand{\mciHeldC}{23}
\newcommand{\mciHeldO}{30}

\newcommand{\mciP}{0.008}

\newcommand{\mldN}{288}

\newcommand{\mldPHeldO}{30}
\newcommand{\mldPN}{48}
\newcommand{\mldPP}{10^{-8}}

\newcommand{\msChecks}{24}

\newcommand{\msMaxDev}{0.05}
\newcommand{\msN}{1200}

\newcommand{\nzFastH}{8}
\newcommand{\nzMidH}{4}
\newcommand{\nzN}{24}
\newcommand{\nzSlowH}{23}
\newcommand{\nzTwoH}{24}
\newcommand{\nzWhiteH}{18}

\newcommand{\omegaG}{3.79}

\newcommand{\parMax}{7.9}
\newcommand{\parMaxLat}{35}
\newcommand{\parSlow}{1.1}
\newcommand{\parSlowLat}{105}
\newcommand{\pctTen}{3.7}

\begin{document}
\title{Hold Before You Step: The Information Requirement of In-Place Push Recovery on Force-Sensorless Humanoids}
\author{Jeonghyeon Kim$^{1}$, Jaesang Kim$^{2}$, and Shinsuk Park$^{1,\ast}$
\thanks{$^{1}$Jeonghyeon Kim and Shinsuk Park are with the Department of Mechanical Engineering, Korea University, Seoul, Republic of Korea (e-mail: ahdtld54@korea.ac.kr; drsspark@korea.ac.kr).}
\thanks{$^{2}$Jaesang Kim is with the Department of Mechanical Engineering, University College London, London, U.K. (e-mail: jaesang8766@gmail.com).}
\thanks{$^{\ast}$Corresponding author: Shinsuk Park.}}
\markboth{Preprint}{Kim \MakeLowercase{\textit{et al.}}: Hold Before You Step}
\maketitle

\begin{abstract}
In-place push recovery preserves a humanoid's task stance. Without force/torque sensors, the controller must decide whether to hold using a delayed and noisy estimate, but how much delay and error it can tolerate is unclear. We measure this information requirement on a simulated Unitree G1 by feeding a fixed hold-or-step controller the true push force, delayed and corrupted. In boundary tests, the initial loss of holdable force agreed with the pendulum rate of about \pctTen\,\% per 10\,ms; actuation delay drew on the same budget, and slow errors of 30\,N were tolerated while 20\,N errors on the push's time scale were not. The requirement guides an estimate that resolves the pelvis inertial measurement unit through the floating-base model, retains the joint accelerations and needs no upper-body torques; for three of four estimators, hold-entry latency predicted the normalised holdable force within 0.02 (registered tolerance 0.08). The resulting controller met its pre-registered recovery criterion against a stepping baseline with \ffCutB\,\% fewer steps and, on holdable pushes that both controllers recovered from, \costABDisp\,\% less base displacement.  On the real G1, the estimate followed a load cell by \hwLatOurs\,ms, against \hwLatMob\,ms for a momentum observer, and with steps disabled, the largest lateral push held per block had a median of \hwAblScoreOurs\,N with the hold layer, whereas without it no block held a push (\hwAblBlocks\ blocks each).\ Disturbance estimates for balance should therefore be judged by the recovery they preserve.
\end{abstract}


\begin{figure}[!t]\centering
\includegraphics[width=\columnwidth]{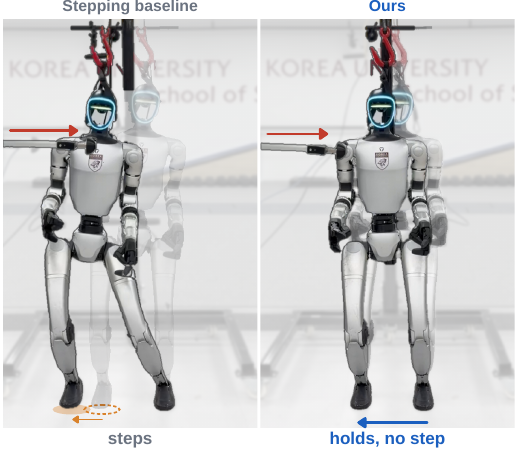}
\caption{A lateral push on the real G1: the stepping baseline steps, whereas our controller holds in place. Illustrative hand pushes in a safety gantry (not a registered evaluation; push forces not matched); poses 0 and 0.6\,s (ours also 1.6\,s) after contact, faint to solid; blue arrow: schematic ground reaction; quantitative results are in Sec.~\ref{sec:exp}.}\label{fig:teaser}
\end{figure}

\section{Introduction}
\IEEEPARstart{H}{umanoid} robots must recover from disturbances while limiting disruption to their ongoing tasks. Recovery steps can interrupt manipulation performed from a fixed stance~\cite{zhang2025tase}, and in human balance, effort favours in-place strategies when they suffice~\cite{afschrift2016}. Recovery should therefore exploit the existing foot contacts and step only when they are insufficient (Fig.~\ref{fig:teaser}).

Existing controllers combine ankle and hip strategies with stepping, or switch between standing and stepping policies~\cite{zhang2025tase,kim2023cpmpc}. Ankle action shifts the centre of pressure (CoP) and hip action regulates centroidal angular momentum within the current contacts, whereas stepping changes the support configuration~\cite{kim2023cpmpc,horak1986}. Capturability analysis characterises whether a given state can be brought to rest with a specified number of steps~\cite{koolen2012}. However, corrective CoP motion is bounded by the support region, so a delayed restoring response can allow motion caused by the disturbance to grow until a new contact becomes necessary.

Without dedicated force/torque (F/T) sensors, observers infer external loads from motion and actuation signals~\cite{flacco2016}, and must also separate a transient push from sustained loads such as a carried object~\cite{song2026mass}. When these estimates determine when and how a controller responds, filtering can suppress noise at the cost of delaying the restoring action~\cite{sun2026tim}, while estimation errors can alter the decision to hold or step. The choice of estimator matters: on the same controller, a tuned momentum observer held a third as many pushes in place as our estimate (Sec.~\ref{sec:closed}). Yet neither a state's capturability nor an estimate's accuracy alone establishes the delay and error that a specified controller can tolerate while retaining recovery without stepping.

\begin{figure*}[t]\centering
\includegraphics[width=\textwidth]{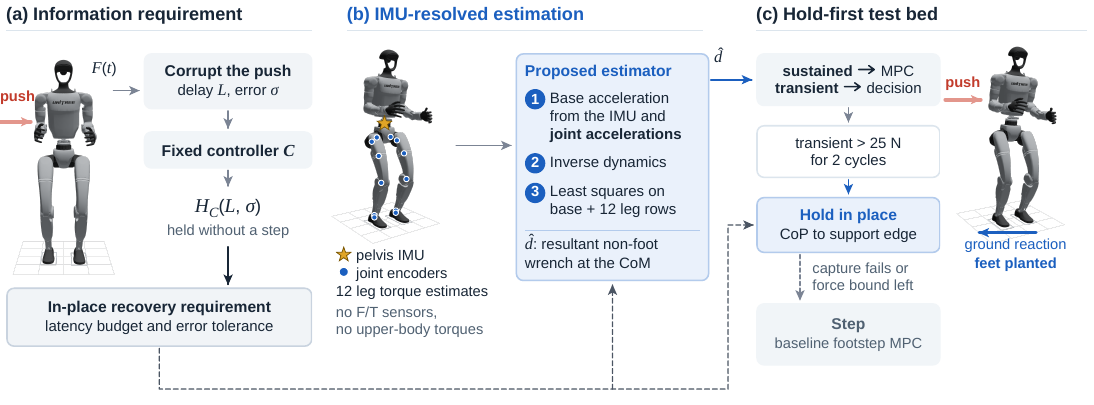}
\caption{Overview. (a) A fixed hold-or-step controller receives the true push force $F(t)$, delayed and corrupted. (b) The estimate $\hat d$ resolves the pelvis IMU through the floating-base model. (c) The hold-first test bed: the sustained part of $\hat d$ (0.5\,s filter) loads the MPC; the transient part drives the hold decision. Dashed: the requirement guides the estimator and the signal paths.}\label{fig:method}
\end{figure*}

We quantify this information requirement on a simulated Unitree G1 by supplying a fixed hold-or-step controller with delayed and corrupted true force. We then use the measured requirement to evaluate estimation and guide the controller's feedback paths. Our contributions are as follows (Fig.~\ref{fig:method}).
\begin{enumerate}
\item \textbf{The information requirement of in-place recovery.} We quantify how information delay and estimation error affect the maximum push force a fixed controller can reject without stepping. The initial loss of holdable force follows the rate predicted by an idealised linear inverted pendulum model. Information and actuation delays share a latency budget, while the tolerated error magnitude depends on the error's time scale.
\item \textbf{IMU-resolved external wrench estimation.} We reconstruct the resultant external force and moment, excluding foot--ground reactions, using a pelvis inertial measurement unit (IMU), joint encoders, and 12 leg torque estimates. Under stated model and contact assumptions, the floating-base reconstruction requires neither F/T sensors nor upper-body torque measurements. It retains joint acceleration terms and avoids differentiating reconstructed base velocity, reducing noise at matched latency relative to the tested momentum observers.
\item \textbf{Disturbance feedback for hold-or-step control.} We use the measured delay and error sensitivity to configure fast decision paths and filtered feedback for force-bound checks and CoP feedforward. In paired simulations, the controller meets the pre-registered recovery criterion against a stepping model predictive control (MPC) baseline with fewer steps and more pushes held in place. On holdable pushes recovered by both controllers, it also reduces base displacement and leg mechanical work.\ On the real G1, with steps disabled, the largest lateral push held per block had a median of \hwAblScoreOurs\,N with the hold layer, whereas without it no block held a push.
\end{enumerate}
Recovery comparisons are conducted in simulation; hardware tests evaluate the estimators against a load cell and during the robot's own arm motion, and the hold layer against the same controller without it.

\section{Related Work}
\emph{Recovery strategies and the step decision.} Capture-point model predictive control coordinates ankle, hip and stepping strategies~\cite{kim2023cpmpc}, and a hierarchical scheme combines a learned standing policy and ankle modulation with a switch to stepping when instability is predicted~\cite{zhang2025tase}. State-based criteria for a necessary step include, besides capturability~\cite{koolen2012}, the extrapolated centre of mass~\cite{hof2005}, the divergent component of motion (DCM)~\cite{englsberger2015} and geometric sets of feasible centroidal moment pivots~\cite{schuller2026}. Standing controllers reject pushes without a step~\cite{castano2016}; steps are triggered by planner feasibility~\cite{khadiv2020,smaldone2021}, predicted instability~\cite{li2025gait} or capture-point and velocity thresholds~\cite{meng2025vicmpc}. Learned policies recover from large pushes~\cite{falcon2025} without an explicit criterion for holding in place. These methods organise recovery from the measured state or an assumed disturbance; we ask what disturbance information the decision needs.

\emph{Timing and uncertainty in balance.} Delayed state feedback stabilises the inverted pendulum only below a critical delay~\cite{milton2009}, and acceleration feedback counteracts reflex delay~\cite{insperger2013}. Bounded sensor and actuator uncertainties yield tracking-error bounds and sampling guidance for legged balance~\cite{villa2019}. Push detectors trade lead time against false alarms~\cite{subburaman2023}, and an earlier recovery step improved push recovery~\cite{li2025gait}. Balance assistance must react faster than physiological responses to help~\cite{beck2023}, and information-theoretic bounds limit sensor-based control~\cite{majumdar2023limits}. These concern the whole loop or human assistance; here only the push information is late or wrong, and, to our knowledge, no study varies its latency and error independently on a fixed controller and measures the in-place recovery that remains.

\emph{Disturbance estimation without F/T sensors.} Feeding an external-force estimate back into a whole-body controller improved push recovery~\cite{li2022eforce}. Generalised-momentum observers estimate external torques without joint accelerations~\cite{deluca2005,haddadin2017} and have been applied to humanoids with all joint torques~\cite{flacco2016}; enhanced observers add a lead term and a Kalman filter to trade phase lag against noise~\cite{sun2026tim}, and learned uncertainty torques, trained with foot F/T data, remove much of the model error~\cite{lim2025}. Other estimators use measured contact forces with an IMU~\cite{benallegue2018,morlando2022}, or resolve the IMU through a centroidal model without the joint accelerations~\cite{fedsi2026}, which carry the inertial force of moving arms. These estimators are evaluated by accuracy and response speed or by closed-loop recovery~\cite{morlando2021}, not against the latency and error that an in-place hold tolerates. We keep the joint accelerations, show that the base and leg rows suffice without foot F/T sensors or upper-body torques (Proposition~\ref{prop:suff}), and evaluate the estimators against that requirement.

\section{The Information Requirement of an In-Place Hold}\label{sec:problem}
\begin{definition}[Information-limited holdability]
Fix a controller $C$, an initial state, a contact set and actuator limits. Let $C$ receive the push information delayed by $L$ and corrupted by an error of standard deviation $\sigma$ (white unless its spectrum is stated). The information-limited holdability $H_C(L,\sigma)$ is the fraction of a fixed set of pushes that $C$ recovers without a step.
\end{definition}
$H_C(0,0)$ is an experimental reference for what the controller holds with ideal information, not a physical viability bound. We also measure the \emph{holdable force}: the first downward crossing of a 50\,\% in-place recovery rate on a force grid, linearly interpolated between the adjacent forces (2.5 or 5\,N spacing; three or four seeds per force). Boundaries outside the tested grid are censored and excluded from comparisons.

\begin{proposition}[Holdable push versus hold-start latency]\label{prop:latency}
Consider the linear inverted pendulum in the ground frame, $\dot\xi=\omega(\xi-p)+F/(m\omega)$, where the DCM $\xi$ and the CoP $p$ are measured from the stance centre and the support $P$ is convex, closed and contains the stance centre ($P=[-b,b]$ along one direction). Let $\xi(0)=\xi_0$ and let a push $F(t)$ act on $[0,T]$, with the DCM displacement
\begin{equation*}
\delta=\frac{1}{m\omega}\int_0^T e^{-\omega s}F(s)\,ds .
\end{equation*}
The CoP stays at $p=0$ until the hold starts at $L\ge0$ and is free in $P$ afterwards. Within this model, a CoP schedule exists after which the state is 0-step capturable~\cite{koolen2012} iff
\begin{equation}
\xi_0+\delta\in e^{-\omega L}P .\label{eq:hold}
\end{equation}
Hence, for $\xi_0=0$, the set of holdable pushes shrinks by the factor $e^{-\omega L}$ in every direction, for any push profile and duration. Along one direction, a constant push $F$ is holdable iff
\begin{equation}
|F|\le F_{\max}(L)=m\omega^2 b\,\frac{e^{\omega T}}{e^{\omega T}-1}\,e^{-\omega L}.\label{eq:fmax}
\end{equation}
\end{proposition}
\begin{proof}
Let $\tau=\max(T,L)$. Variation of constants gives
\begin{equation*}
\xi(\tau)=e^{\omega\tau}(\xi_0+\delta)-\int_L^\tau\omega e^{\omega(\tau-s)}p(s)\,ds .
\end{equation*}
The weights of the integral sum to $e^{\omega(\tau-L)}-1$, so, since $P$ is convex and closed, the integral over all schedules in $P$ takes exactly the values in $(e^{\omega(\tau-L)}-1)P$ (constant schedules attain each of them). Some schedule therefore ends in $P$ iff $e^{\omega\tau}(\xi_0+\delta)\in P+(e^{\omega(\tau-L)}-1)P=e^{\omega(\tau-L)}P$, using $\lambda P+\mu P=(\lambda+\mu)P$ for convex $P$ and $\lambda,\mu\ge0$; this is \eqref{eq:hold}. If $\xi(\tau)\in P$, the CoP $p=\xi(\tau)$ keeps $\dot\xi=0$, so the robot stops without a step. For a constant push, $\delta=F(1-e^{-\omega T})/(m\omega^2)$, which gives \eqref{eq:fmax}.
\end{proof}
The proposition extends 0-step capturability~\cite{koolen2012,schuller2026} to late push information. On the G1 ($\omega=\omegaG$\,s$^{-1}$), each 10\,ms of latency costs about \pctTen\,\% of the holdable force.

The model ignores angular momentum and foot rotation, and stance feedback before the hold shifts $\xi_0$, so $e^{-\omega L}$ is exact only without pre-hold stance feedback; we therefore test the ratio, not the absolute boundary, on constant 0.2\,s pushes.

Two consequences follow within the model. First, a command delay $D_a$ that postpones the CoP schedule replaces $L$ by $L+D_a$: information and actuation delay share one budget. Second, a hand load or own arm motion that reserves part of the support restricts the CoP to a convex, closed $P_r\subseteq P$ containing the stance centre, for which the proof holds unchanged; along one direction, with $P_r=[-(b-b_r),b-b_r]$ and $\xi_0=0$,
\begin{equation}
|\delta|\le e^{-\omega L}\,(b-b_r),\label{eq:budget}
\end{equation}
so the delay factor multiplies whatever margin remains.

\section{An Estimate That Meets the Requirement}\label{sec:est}
The requirement asks for an estimate that is fast at low noise. A momentum observer differentiates, in effect, a base velocity reconstructed from leg kinematics; we instead measure the base acceleration with the pelvis IMU and resolve it through the model. The floating-base dynamics are
\begin{equation}
M(q)\ddot q+h(q,\dot q)=S^\top\tau+\textstyle\sum_i J_i^\top f_i+\sum_k J_k^\top w_k ,\label{eq:dyn}
\end{equation}
with foot wrenches $f_i$ and non-foot external wrenches $w_k$. We partition $q=(q_b,q_L,q_U)$ into the floating base (the pelvis), the 12 leg joints and the 17 upper-body joints. Let $d$ be the resultant wrench of all $w_k$, expressed at the centre of mass (CoM).

\emph{Estimate.} Each control cycle has three steps. First, the pelvis IMU's specific force $a_s$ and differenced angular rate are affine in the base acceleration for given joint accelerations, $[\dot\omega_s;a_s]=J_s\ddot q+\dot J_s\dot q-[0;R_s^\top g]$. We solve these six rows for the base acceleration, with the joint accelerations from control-cycle differences of the estimated joint velocities. Second, we evaluate the inverse dynamics. Third, we solve the 6 base and 12 leg rows of \eqref{eq:dyn} by unweighted, unregularised least squares for $d$ (at the CoM) and the 6D wrench of each foot in contact (at the ankle), using the leg torque estimates $\tau_L$ with the model's leg friction subtracted. A first-order filter of bandwidth $K=100$\,s$^{-1}$, $y_k=y_{k-1}+\min(K\Delta t,1)\,(x_k-y_{k-1})$, smooths the result. The estimate uses no upper-body torque, but the positions and differenced velocities of all joints. It relies on current-based leg torques, whose errors under dynamic loading remain to be characterised.

\begin{proposition}[Lower-body sufficiency]\label{prop:suff}
Assume (A1) the non-foot wrenches act on the pelvis or the upper body, not on the legs; (A2) each leg in contact has six joints and a non-singular foot Jacobian $J_{i,L}$, and each foot wrench is a full 6D wrench; (A3) the generalised position, velocity and acceleration of all joints and the base are available, and the model is exact. Then the base and leg rows of \eqref{eq:dyn} determine the foot wrenches and the resultant $d$ uniquely, and $\tau_U$ does not enter. Only the resultant, not the location or distribution of the non-foot wrenches, is identifiable.
\end{proposition}
\begin{proof}
In a kinematic tree the row of joint $j$ in $J_k^\top$ is non-zero only if body $k$ is distal to $j$. Feet are distal only to their own leg joints. By (A1), the leg rows contain no non-foot wrench and read $M_L\ddot q+h_L=\tau_L+J_{i,L}^\top f_i$ per leg, and by (A2) $J_{i,L}$ is invertible, which fixes $f_i$. The base rows carry no actuation and are the whole-body Newton--Euler equations $\dot h_G=\sum_i\mathrm{Ad}_i^\top f_i+d+mg$ about the CoM, with the centroidal momentum $h_G$, the transforms $\mathrm{Ad}_i^\top$ of the foot wrenches to the CoM and the gravity wrench $mg$; they fix $d$.
\end{proof}
Upper-body model and state errors still enter through $M\ddot q$ and $h$, and (A2) fails near a leg singularity (condition number below 70 in our runs).

\begin{proposition}[Noise scaling of the two structures, first-order approximation]\label{prop:pareto}
Model the base-velocity and joint-velocity errors as zero-mean noise with standard deviations $\sigma_{v_b}$ and $\sigma_{\dot q}$ whose power lies mostly at frequencies well above $K$, and the accelerometer error as white noise of density $\sigma_a$. Pass each raw estimate through a first-order filter of bandwidth $K$ and neglect sampling ($K\Delta t\ll1$). For a fixed disturbance amplitude and detection threshold, the detection latency of a step disturbance then scales with $1/K$. (i) The noise of a floating-base momentum observer has standard deviation $\approx K\,m\,\sigma_{v_b}$, with the total mass $m$. (ii) The noise of the IMU-resolved estimate has a term $\propto\sqrt K\,m\,\sigma_a$ and, for each joint, a term $\propto K\,m_{\mathrm{dist}}\,\sigma_{\dot q}$, with the mass $m_{\mathrm{dist}}$ distal to that joint.
\end{proposition}
\begin{proof}
With the generalised momentum $\pi=M\dot q$ and $\dot\pi=\tau+\beta+\tau_{\mathrm{ext}}$, the observer residual $r=K(s\pi-\beta-\tau)/(s+K)$ passes a momentum error as $Ks\,\delta\pi/(s+K)$, a high-pass of gain $K$ that transmits nearly all the variance of a velocity error with power well above $K$, giving $\approx K\,m\,\sigma_{v_b}$. In the IMU-resolved estimate, white accelerometer noise passes through $K/(s+K)$ with standard deviation $\sigma_a\sqrt{K/2}$, and differenced joint velocities enter through the momentum-Jacobian columns, which scale with the distal mass, and the same high-pass, giving $\approx K\,m_{\mathrm{dist}}\,\sigma_{\dot q}$.
\end{proof}
Where the base-velocity error dominates, these relations predict a noise advantage at equal latency that grows as latency shrinks, the range Proposition~\ref{prop:latency} singles out; the joint-velocity term or sampling ($K\Delta t=0.5$ at 200\,Hz) can dominate instead, so Section~\ref{sec:expEst} measures noise at matched latency.

\section{Test Bed: A Hold-First Controller}\label{sec:control}
The test bed isolates the hold-or-step decision: an estimate-driven hold layer is added to a stepping baseline, which consists of a centroidal model predictive controller (MPC), a footstep MPC that chooses the step location and swing duration, and a whole-body inverse-dynamics quadratic program (QP). The estimate is split into a sustained part (0.5\,s filter), which the baseline treats as a load, and a transient part $d_{\mathrm{fast}}$. When the horizontal force of $d_{\mathrm{fast}}$ exceeds 25\,N for two control cycles, the controller enters the hold mode, an ankle-priority mode, before the planner reacts. In this mode, a linear program over the QP's variables drives the CoP to the support edge to cancel the transient; it limits the centroidal angular-momentum rate and maintains at least 10\,\% of the weight on each foot. The hold is thus designed to realise the CoP schedule of Proposition~\ref{prop:latency} after the hold-entry latency $L$, which includes the two confirmation cycles and is measured in Sec.~\ref{sec:expEst}. Steps are blocked during a hold. The hold ends and a step is triggered when a 0-step capture prediction fails (linear inverted pendulum over 0.3\,s, CoP at the support edge toward the DCM) or when the transient leaves the force bound, an ellipse of 60\,N fore-aft and 75\,N lateral (45\,N fore-aft while the arms are commanded to move). All constants were fixed on development seeds before the evaluations.

\emph{Two configurations.} This \emph{evaluation configuration} measures the requirement and serves the offline comparison; the \emph{final configuration} (Sec.~\ref{sec:expLat}), used for the closed-loop estimator boundaries and all controller comparisons, changes the information paths and, during arm motion, the hold reference.

\section{Experiments}\label{sec:exp}
\begin{figure*}[t]\centering
\includegraphics[width=\textwidth]{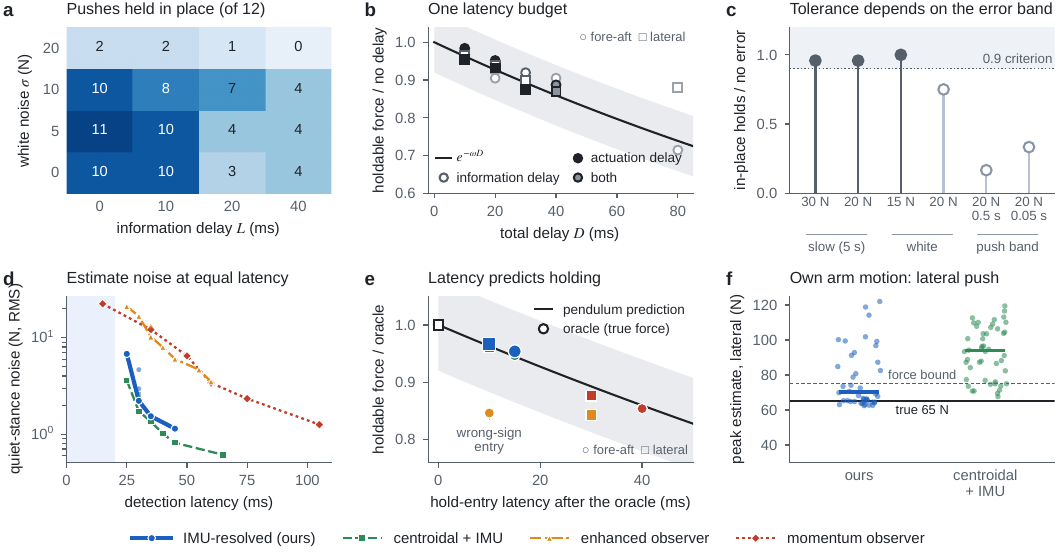}
\caption{The information requirement of the hold (top) and the estimators against it (bottom). (a) Pushes held without a step when the true push force reaches the hold decision delayed by $L$ and with white noise $\sigma$ (\lnN\ near-boundary pushes per cell; 95\,\% Wilson intervals at $\sigma=0$: \lnZeroCI\,\% at 0\,ms, \lnTwentyCI\,\% at 20\,ms). (b) Holdable force relative to no delay versus the combined actuation and information delay $D$ (608 runs) and, hollow markers, versus information delay with the capture prediction as the only step trigger (boundary search); line: $e^{-\omega D}$, band $\pm0.08$. (c) Holds under injected errors of different correlation times, relative to no error (240 runs); filled: within the registered tolerance of 0.9 (shaded). (d) Quiet-stance noise versus detection latency over each estimator family's tuning range (55 logs; lines: lowest noise at each latency, pale points: all settings). (e) Closed-loop holdable force against the prediction from hold-entry latency (380 runs); circles fore-aft, squares lateral. (f) Own arm motion, post hoc: peak estimate of each 65\,N lateral boundary push (dots; bar: median), ours against the centroidal estimate; dashed: lateral force bound. Shading: $\pm0.08$ tolerance in (b) and (e), first 20\,ms in (d).}\label{fig:necessity}
\end{figure*}
\emph{Setup.} A Unitree G1 with 29 degrees of freedom is simulated in MuJoCo~\cite{todorov2012} at 200\,Hz with encoder, torque-estimate and IMU noise, a 10\,ms observation delay and the IMU at the pelvis, as on the hardware; pushes of 0.2\,s act on the torso. A run \emph{fails} if a body other than the feet touches the ground, the pelvis drops below 0.5\,m or a swing foot does not touch down (a tilt stop at 1.0\,rad terminates divergent runs); it is \emph{recovered} if it does not fail and ends standing (last 0.5\,s: pelvis above 0.6\,m, mean speed below 0.1\,m/s), and \emph{held in place} if recovered without a step.

\emph{Protocol.} Every comparison was pre-registered with a recorded hash before its runs on new seeds (offline comparisons: before the analysis), and no decision threshold was changed afterwards, except for the outcome-definition amendment below; analyses marked post hoc were not registered. A registered amendment replaced the original 0.35\,rad tilt stop, which also ended standing runs during arm lifts, and re-scored all \fcN\ runs (\fcUp\ to recovered, \fcDown\ to failed, two decisions changed; counts were stable for tilt stops of 0.6--1.2\,rad). One-sided tests in the registered direction are exact McNemar (paired recovery, same condition and seed), Wilcoxon signed-rank (paired estimator metrics, steps) and Mann--Whitney (hardware blocks). The $p$-values are nominal and unadjusted except for the Holm-corrected E2 follow-up; runs sharing a seed share sensor noise and torque-estimate biases (four or six seeds per cell). Registrations, analysis code and run summaries are available from the authors on request; project page and videos: \repoURL.

\emph{Timing.} The true force bypasses the 10\,ms observation delay ($L=0$). \emph{Hold-entry latency} runs from push onset to the first hold cycle; offline \emph{detection latency} ends when the filtered estimate ($K=100$\,s$^{-1}$) exceeds 25\,N on two consecutive samples.

\subsection{The information requirement}\label{sec:expA}
\emph{Near the boundary, holding is lost within the first tens of milliseconds.} We fixed the controller and replaced the estimate in its whole hold decision (entry, ankle target, capture prediction and force bound) by the true push force, delayed by $L$ and corrupted by white noise $\sigma$ per axis, on pushes near the boundary ($\pm70$\,N lateral, $\pm50$\,N fore-aft and diagonal; Fig.~\ref{fig:necessity}a). With ideal information, 10 of 12 pushes were held; a delay of 10\,ms left 10 and one of 20\,ms left 3. Neither registered monotonicity prediction held over the full grid: 20\,N of noise suppressed holding even without delay, whereas moderate noise at 20\,ms raised the holds (3 to 4 and 7) by triggering entry earlier. Our estimate held 7 of 12.\

\emph{The loss follows the pendulum rate.} With the capture prediction as the only step trigger, a boundary search (Fig.~\ref{fig:necessity}b, light markers) found the normalised holdable force within $\pm0.08$ of $e^{-\omega L}$ (about one 5\,N step of the force grid) at 20 and 40\,ms in both directions and at 80\,ms fore-aft. Laterally, the boundary stopped falling after 40\,ms, when the stance feedback on the wider support became dominant. The four tested delays support only the initial rate.

\emph{Actuation delay draws on the same budget.} Delaying the joint commands to the drives by 10--30\,ms cost as much holdable force as an equal information delay (largest difference \latMaxDiff\ of the no-delay boundary; tolerance 0.05), and every combination followed the rate of the total delay within $\pm\latMaxDev$ (608 runs; tolerance $\pm0.08$; Fig.~\ref{fig:necessity}b).

\emph{The rate holds under hand loads and own arm motion.} With a 2\,kg hand load, arm motion or both (\msN\ runs), the normalised holdable force matched $e^{-\omega L}$ in all \msChecks\ checks (largest deviation \msMaxDev), supporting the ratio in \eqref{eq:budget}.

\emph{Which errors matter depends on their time scale.} We added Ornstein--Uhlenbeck errors of equal stationary standard deviation but different correlation times to the true force from the run start (240 runs; Fig.~\ref{fig:necessity}c). Slow errors of 20 and 30\,N (correlation time 5\,s) left \nzSlowH\ of \nzN\ near-boundary pushes held in place (\nzN\ without error), because the hold entry acts on the estimate minus its sustained part; an error of 20\,N on the time scale of a push (0.5 and 0.05\,s) left \nzMidH\ and \nzFastH. White noise was tolerated at 15\,N (\nzTwoH\ of \nzN; registered after the 20\,N result) but not at 20\,N (\nzWhiteH), where it kept the hold mode active in quiet stance. This brackets this controller's tolerance for white noise only, excluding motion-induced and endogenous errors.

\subsection{Estimators against the requirement}\label{sec:expEst}
We compared ours with three alternatives: a momentum observer using all 29 joint torques~\cite{deluca2005,haddadin2017} (gain 30\,s$^{-1}$ with a 0.1\,s output filter, tuned in closed loop), our re-implementation of the enhanced momentum observer with a Kalman filter~\cite{sun2026tim} (measurement noise matched to ours), and a centroidal estimate that is also IMU-resolved but omits joint accelerations~\cite{fedsi2026}.

\emph{Offline, ours is faster at lower noise, consistent with Proposition~\ref{prop:pareto}.} On the same pelvis-IMU logs of 55 pushes (30--70\,N along $x$, $y$, $xy$), ours detected pushes after \ecThreeLatI\,ms on average against \ecThreeLatMob\,ms for both momentum observers, with the lowest push-phase error at every magnitude and direction (root-mean-square, RMS, \ecThreeRmsI\ against \ecThreeRmsLitMin--\ecThreeRmsLitMax\,N; paired Wilcoxon $p<0.002$). Across the tuning sweeps (Fig.~\ref{fig:necessity}d), the observer-to-ours noise ratio at matched latency rose from \parSlow\ at \parSlowLat\,ms to \parMax\ at \parMaxLat\,ms. The ranking held at twice the sensor noise; at the highest level, ours and the centroidal estimate did not differ significantly (\ecSevI\ against \ecSevC\,N, $p=\ecSevP$).

\emph{In closed loop, latency predicted holding.} Proposition~\ref{prop:latency} predicts a normalised holdable force of approximately $e^{-\omega\Delta L}$ from the hold-entry latency $\Delta L$ relative to the true-force oracle, provided the estimation error remains within tolerance. In the final configuration (380 runs; Fig.~\ref{fig:necessity}e), ours entered after \estOursLatL--\estOursFaL\,ms and kept \estOursFaR--\estOursLatR\ of the true-force boundary, the tuned momentum observer entered after \estMobLatL--\estMobFaL\,ms and kept \estMobFaR--\estMobLatR, and both matched the prediction within \estMaxDev\ (tolerance 0.08), as did the centroidal estimate (\estCimuFaR--\estCimuLatR). The prediction, $\omega$ and the tolerance were fixed in the registrations; nothing was fitted to the estimator boundaries. Fore-aft, the enhanced observer reached \estEmoFaR\ instead of the predicted \estEmoFaPred. A wrong-sign transient of 21--32\,N, about 10\,ms after each onset, started holds in the wrong direction. Post hoc bootstrap 95\,\% intervals of the deviation (runs resampled within force levels, boundaries and latencies recomputed; also with seeds as clusters) lay within $\pm0.05$ for the other three estimators and at $-0.15$ to $-0.02$ for the enhanced observer, excluding the force-grid spacing and the boundary interpolation.

\emph{Joint accelerations matter when the robot moves its arms.} The two IMU-resolved estimates differ only in the joint-acceleration terms, which carry the inertial force of the arms. At the in-place boundary during four motion-capture arm motions ($\pm40$\,N fore-aft, $\pm65$\,N lateral; \mcN\ pairs), ours held \mcHeldO\ of the \mcN\ pushes in place against \mcHeldC\ ($p<10^{-7}$), recovered \mcRecO\ against \mcRecC\ runs and took \mcStO\ against \mcStC\ steps. Post hoc inspection showed that the centroidal method overestimated the 65\,N lateral pushes (median peak \mcPkC\ against \mcPkO\,N; Fig.~\ref{fig:necessity}f), crossing the force bound and releasing the hold. With 50\,N pushes, ours held \mciHeldO\ of 48 in place against \mciHeldC\ ($p=\mciP$), under the amended outcome definition.

\subsection{Requirement-guided design}\label{sec:expLat}
Motivated by the measured sensitivity to delay, the paths are treated differently. The \emph{decision} (hold entry, capture prediction) uses the unfiltered estimate and a one-cycle entry instead of the filter and two-cycle confirmation; the \emph{force bound} uses a five-cycle median that rejects brief IMU spikes (up to 96\,N during a 60\,N push) that otherwise triggered unnecessary steps; the \emph{actuation} (CoP feedforward) keeps the filter, because it closes a loop through the accelerations the estimate uses and oscillated at the Nyquist frequency without it. Without filtering on any path, the controller held more single pushes (\latZeroF\ to \latZeroL, $p=\latP$) but recovered fewer successive pushes and stepped more during arm motion. During commanded arm motion, the hold also keeps the MPC's planned angular-momentum rate and vertical motion as its reference. Against the evaluation configuration (104 runs each), the final configuration raised single-push holds from \lthZeroM\ to \lthZeroL\ ($p=\lthP$) and cut the steps by \lthStCut\,\%; the comparison evaluates these changes together.

\subsection{Final comparison}\label{sec:closed}
\begin{table}[t]\centering\footnotesize
\caption{Final comparison (\ffN\ paired runs: 50--80\,N single and successive pushes). All rows share the plant, stance controller, footstep MPC and pushes. Baseline: feasibility trigger with step-timing MPC. Literature rules: our implementations on our hold layer, not the published controllers; learned policies~\cite{falcon2025} were not compared. Zero-step: recovered without a step.}\label{tab:main}
\begin{tabular}{lrrr}
\toprule
Controller & Recovered & Zero-step & Steps\\
\midrule
\textbf{Ours} & \textbf{\ffH} & \textbf{\ffZ} & \textbf{\ffSt}\\
Stepping baseline & \ffHB & \ffZB & \ffStB\\
Predicted instability~\cite{li2025gait} & \ffHLi & \ffZLi & \ffStLi\\
Capture point + velocity~\cite{meng2025vicmpc} & \ffHMeng & \ffZMeng & \ffStMeng\\
0-step capture gate~\cite{pratt2006,koolen2012} & \ffHCg & \ffZCg & \ffStCg\\
Ours, momentum observer & \ffHMob & \ffZMob & \ffStMob\\
\bottomrule
\end{tabular}
\end{table}
\emph{The recovery criterion was met with half the steps} (Table~\ref{tab:main}). The registered criteria allowed at most 4 fewer recovered runs of the 80 and at most 80\,\% of the steps (pass criteria, not a non-inferiority test). Against the stepping baseline, ours recovered \ffH\ against \ffHB\ runs, held \ffZ\ pushes in place against none and took \ffCutB\,\% fewer steps. Because failed runs end at different times, we also compared steps on the runs both recovered (post hoc): \jrSOB\ against \jrSBB\ ($p\jrPB$). The gain lies in single pushes (\ffSingO\ against \ffSingB\ of \ffSingN); on the \ffSuccN\ successive-push runs, no controller held in place, and ours recovered \ffSuccO\ against \ffSuccB.

\emph{Both the step decision and its information matter.} With the step-triggering rules from the literature on the same hold layer, ours recovered more runs than the two capture-point rules ($p\le0.008$) and, against the predicted-instability rule, met the recovery criterion with \ffCutLi\,\% fewer steps and more than twice the pushes held in place. With the tuned momentum observer in place of our estimate, the same controller met the recovery criterion but took \ffStMob\ instead of \ffSt\ steps and held \ffZMob\ instead of \ffZ\ pushes in place.

\emph{The advantage persists during the robot's own arm motion and under hand loads.} Over \ffCondN\ such runs, ours recovered \ffCondH\ against \ffCondHB, taking \ffCondSt\ instead of \ffCondStB\ steps. With a 2\,kg hand load during arm motion (\mldN\ runs), ours held \mldPHeldO\ of the \mldPN\ runs with a 50\,N push in place, against one for the baseline ($p<\mldPP$).

\begin{figure*}[t]\centering
\includegraphics[width=\textwidth]{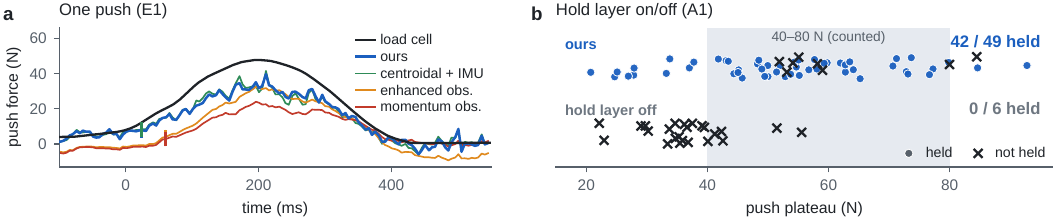}
\caption{Hardware, registered tests. (a) E1: one push from behind, load-cell force (10\,ms mean) and the four estimates along the push direction, replayed causally at 200\,Hz; ticks: each signal's 50\,\% crossing of its own plateau. Ours and the centroidal estimate nearly coincide on a push; they differ in the joint-acceleration terms, which mattered during arm motion in simulation (Fig.~\ref{fig:necessity}f) and in the delay-tuned hardware test (E2). (b) A1, steps disabled: every counted lateral push by its load-cell plateau, held (dots) or not (crosses), for ours and the same controller with the hold layer off; counts: held of the 40--80\,N pushes (shaded).}\label{fig:hw}
\end{figure*}
\subsection{Cost of recovery}\label{sec:cost}
Holding in place is also cheaper. A registered comparison (348 runs) measured costs over 2.9--6.0\,s (push at 3\,s) on runs both recovered, as median paired reductions (thresholds 50, 15 and 25\,\%). On holdable pushes (30--50\,N; \costAN\ runs), ours held \costAOZero\ in place against \costABZero\ for the baseline and reduced the net horizontal base displacement by \costABDisp\,\% (medians \costABDispO\ against \costABDispB\,mm), the peak foot force by \costABPeak\,\% and the leg work $\sum_j\int|\tau_j\dot q_j|\,dt$ over the 12 leg joints by \costABWork\,\%. With hand loads and objects picked and placed between pushes (\costCN\ runs), ours held all \costCOZero\ in place, whereas the baseline recovered \costCBRec\ with \costCBSt\ steps. Because the baseline stepped even at $\pm30$\,N fore-aft, these savings reflect that stepping and concern motion and loading, not task performance.

\begin{table}[t]\centering\footnotesize
\caption{Hardware estimators at their operating settings (E1, medians). Delay: behind the load cell at 50\,\% of the plateau; error, noise: push-phase and quiet-stance RMS.}\label{tab:hw}
\begin{tabular}{lrrr}
\toprule
Estimator & Delay (ms) & Error (N) & Noise (N)\\
\midrule
\textbf{IMU-resolved (ours)} & \textbf{\hwLatOurs} & \hwErrOurs & \hwNoiseOurs\\
Centroidal + IMU & \hwLatCimu & \hwErrCimu & \hwNoiseCimu\\
Enhanced observer & \hwLatEmo & \hwErrEmo & \hwNoiseEmo\\
Momentum observer & \hwLatMob & \hwErrMob & \hwNoiseMob\\
\bottomrule
\end{tabular}
\end{table}

\subsection{Hardware validation}\label{sec:hw}

The G1 stood in a gantry, with the IMU at the pelvis as in simulation. Pushes were applied by hand through a load cell (7.5\,kHz) logged on the robot's clock. For E1 and E2, the learned policy~\cite{falcon2025} kept the robot standing and moved its arms, while our controller computed its commands without sending them. The estimators were replayed causally using the 1\,kHz joint and IMU logs at the 200\,Hz control rate with the code and settings of the simulation comparisons. Metrics, exclusion rules (load cell and foot contact only) and tests were registered after a pilot session; the \hwNPush\ pushes of \hwFmin--\hwFmax\,N came from two sessions, in four directions.

\emph{Detection latency and error (E1).} As hand pushes vary, latency was measured between the load cell's and each estimate's crossing of 50\,\% of its own plateau (gain-normalised). Our estimate responded earlier than the momentum observer (paired difference \hwLatPair\,ms, ours earlier in \hwLatEarlier\ pushes, $p=\hwLatP$; 2 pushes without a crossing excluded; Table~\ref{tab:hw}, Fig.~\ref{fig:hw}a), with a lower push-phase error but a slightly higher quiet-stance noise; at these unmatched operating settings, the matched-latency noise comparison is left to E2. 

\emph{Own arm motion (E2).} During four scripted arm motions without a push, the peak spurious estimate at the registered bandwidth was \hwArmOurs\,N for ours and \hwArmCimu\,N for the centroidal estimate (medians); ours was lower in \hwArmLower\ of \hwArmN\ repetitions ($p=\hwArmP$), so the registered test was not met. Post hoc, the pelvis IMU showed five to ten times its quiet vibration above 15\,Hz, which a faster estimate transmits, so we registered a second test on new recordings (two sessions, \hwArmMatchN\ repetitions), filtering both IMU-resolved estimates at \hwMatchK\,s$^{-1}$, the rate chosen to approximate the observer's E1 delay. The peak was \hwArmMatchOurs\,N for ours against \hwArmMatchCimu\,N for the centroidal estimate (lower in \hwArmMatchLowerCimu\ of \hwArmMatchN) and \hwArmMatchMob\,N for the momentum observer (lower in \hwArmMatchLowerMob\ of \hwArmMatchN, not during the lift: \hwArmLiftMob\ against \hwArmLiftOurs\,N). Both predictions were met after Holm correction (unadjusted $p=\hwArmMatchPCimu$ and $p=\hwArmMatchPMob$). On pushes recorded in the same sessions, ours still responded earlier than the momentum observer (\hwMatchDelayOurs\ against \hwMatchDelayMob\,ms). At this setting, ours read the least spurious force, consistent with Proposition~\ref{prop:pareto}; at its operating bandwidth it trades this margin for an earlier response.

\emph{Holding in place.} With our controller in command and steps disabled, \hwHoldOurs\ of \hwHoldN\ lateral hand pushes of \hwHoldRange\,N (0.25--1.2\,s) were held without a step, under a registered outcome rule (no foot lift, no pelvis tilt change above 8$^\circ$, no straightened knee and no guard stop within 3\,s of onset). The hold mode started a median of \hwEntryLat\,ms after the load-cell force crossed 25\,N (interquartile range \hwEntryIQR\,ms). These pushes are the no-delay arm of a registered latency test that was stopped (Sec.~\ref{sec:concl}). The rounded sole may reduce the usable CoP margin, an effect not quantified here.

\emph{Hold-layer ablation (A1).} A registered test alternated blocks of ours and of the same controller with only the hold entry disabled, both with steps disabled. Each condition had \hwAblBlocks\ blocks in two sessions, each an ascending series of shoulder-height lateral pushes (aimed 30--80\,N) ended by a failure; held as above, with a tilt change below 12$^\circ$. The primary measure was the block score, the largest held plateau (capped at 80\,N; 0 for a block without a held push). Its median was \hwAblScoreOurs\,N for ours and \hwAblScoreAbl\,N for the ablation, which held no counted push in any block (one-sided Mann--Whitney, $p=\hwAblP$, blocks treated as independent; each session alone: 62 and 75\,N against 0\,N; unchanged with the 3\,s window from release instead of onset). Because blocks ended at a failure, the ablation met fewer high forces, so the per-push counts are descriptive: between 40 and 80\,N, ours held \hwAblHeldOurs\ of \hwAblNOurs\ pushes and the ablation \hwAblHeldAbl\ of \hwAblNAbl\ (Fig.~\ref{fig:hw}b). The ablation unloaded a foot in \hwAblLift\ of \hwAblAllNAbl\ pushes (\hwAblFailRange\,N), consistent with the hold preserving contact; ours, keeping 10\,\% on each foot, failed at \hwAblOursFailRange\,N. A registered test with sustained (2\,s) pushes was inconclusive, as push duration depended on the outcome; descriptively, the ablation unloaded a foot in all \hwSusNAbl\ pushes and ours held \hwSusHeldOurs\ of \hwSusNOurs.

\section{Conclusion}\label{sec:concl}
This work quantified, for a fixed controller, the disturbance information needed for recovery without stepping, making recovery capability an estimator design criterion. Controlled simulations revealed a shared latency budget for information and actuation, while errors varying during a push were more disruptive than slow errors. An IMU-resolved wrench estimate gave timely feedback without F/T sensors, with lower noise than momentum observers at matched latency. This requirement predicted, from hold-entry latency alone, the normalised holdable force of three of four estimators without fitting; the fourth showed that the sign of early errors also matters. Guided by this requirement, with fast decision paths and filtered feedback, the controller met its pre-registered recovery criterion with \ffCutB\,\% fewer steps than the stepping baseline. On the real G1, the estimate followed a load cell by \hwLatOurs\,ms against \hwLatMob\,ms for the momentum observer. With stepping disabled, a registered ablation confirmed the hold layer's benefit for holding lateral pushes in place.

A key limitation is that hand-push variability prevented us from resolving the predicted delay effect on hardware, and recovery with stepping was not evaluated there. Future work should test this effect with a repeatable pusher and combine the hold layer and its information requirement with learned stepping policies such as~\cite{falcon2025}. More broadly, stating the information that an in-place hold needs lets estimators and feedback paths be compared by the recovery they preserve, not by accuracy alone.

\bibliographystyle{IEEEtran}
\bibliography{refs}
\end{document}